\PassOptionsToPackage{hyphens}{url}
\documentclass[11pt,letterpaper,logo,onecolumn]{hy_llmeval_style}

\usepackage[authoryear,sort&compress,round]{natbib}
\usepackage[most,breakable,skins]{tcolorbox}
\tcbuselibrary{skins,breakable}
\usepackage{microtype}
\usepackage{amsthm}
\DeclareMathAlphabet{\amsbb}{U}{msb}{m}{n}
\let\mathbb\amsbb
\usepackage{multirow}
\usepackage{float}
\usepackage{placeins}

\let\cite\citep
\hypersetup{
  citecolor=gblue9,
  linkcolor=gblue9,
  urlcolor=gblue9
}

\definecolor{exploitred}{HTML}{991B1B}
\definecolor{repairgreen}{HTML}{065F46}
\definecolor{boxbg}{HTML}{F9FAFB}
\definecolor{boxframe}{HTML}{D1D5DB}

\newcommand{\oer}{\textsc{OER}}
\newcommand{\oerlocal}{\textsc{OER-Local}}
\newcommand{\oerprop}{\textsc{OER-Prop}}
\newcommand{\cotpen}{\textsc{CoT-Penalty}}
\newcommand{\codepen}{\textsc{Code-Penalty}}
\newcommand{\klreg}{\textsc{KL-Reg}}
\newcommand{\execpen}{\textsc{Exec-Penalty}}
\newcommand{\quar}{\textsc{Quarantine}}
\newcommand{\rollmask}{\textsc{Rollout-Mask}}
\newcommand{\auditlb}{\textsc{Audit-LB}}
\newcommand{\offhard}{\textsc{Offline-Harden}}
\newcommand{\offprop}{\textsc{Offline-Prop}}
\newcommand{\autocluster}{\textsc{OER-Auto\-Cluster}}
\newcommand{\reffree}{\textsc{OER-RefFree}}
\newcommand{\midext}{\textsc{Mid-Extinction}}
\newcommand{\auchack}{\ensuremath{\mathrm{AUC}_{\mathrm{hack}}}}

\newtheorem{proposition}{Proposition}
\newcommand{\narrowtable}[1]{{\small\setlength{\tabcolsep}{6.5pt}#1}}
\newcommand{\qualboxwidth}{0.92\textwidth}
\newcommand{\algboxwidth}{0.94\textwidth}

\title{Online Environment Repair for Coding Agents}

\author{
  \textbf{Sheng Lu}, \textbf{Rui Mao}, \textbf{Zewei Chen}, \textbf{Chihua Ying}, \textbf{Beiduo Chen}, \textbf{Dingxin Hu}, \textbf{Tinghao Yu}\textsuperscript{$\dagger$}\\
  \vspace{0.20cm}
  \normalfont\normalsize Hunyuan Team, Tencent\\
  \vspace{0.08cm}
  \normalfont\small\texttt{\{\href{mailto:jonathanslu@tencent.com}{jonathanslu}, \href{mailto:dingxinhu@tencent.com}{dingxinhu}, \href{mailto:maxwellyu@tencent.com}{maxwellyu}\}@tencent.com}
}
\correspondingauthor={Tinghao Yu\textsuperscript{$\dagger$} (\href{mailto:maxwellyu@tencent.com}{maxwellyu@tencent.com})}

\begin{abstract}
Programmatic verifiers provide scalable reward signals for training coding agents with reinforcement learning, yet subtle harness defects and incomplete unit tests allow policies to earn full credit without solving the intended task. Existing training-time defenses intervene either by penalizing suspicious outputs or by filtering flagged tasks and rollouts. However, output penalties leave the underlying verifier exploitable, driving policies to disguise shortcuts under reasoning monitors or zeroing out valid solutions under detector false positives, whereas task and rollout filtering discards or mutes the hardest problems where shortcuts emerge earliest. Pre-training verifier hardening repairs the environment directly, but misses vulnerability classes that surface only as the policy improves during reinforcement learning. In this paper, we propose \emph{online environment repair}, which leaves triggering rollout rewards unchanged to remove hiding incentives and asynchronously patches exploited environments during training. Candidate repairs triggered by execution invariants or differential fuzzing are deployed only after passing three automated validation gates that block the exploit while preserving valid solutions, canceling detector false positives and enabling structural propagation across repositories without scaffold labels or reference solutions. Across 20 controlled GRPO experiments on heterogeneous software-engineering repositories, online environment repair reduces cumulative reward hacking from $36.9\%$ to $0.06\%$ and improves held-out Pass@8 from $21.7\%$ to $46.7\%$, consistently outperforming reward-penalty, rollout-filtering, and static-hardening baselines while eliminating zero-shot exploitation on unseen loophole classes.

\end{abstract}

\begin{document}
\raggedbottom
\fancyhead[C]{\footerfont Online Environment Repair for Coding Agents}

\maketitle

\section{Introduction}

\begin{figure*}[t]
    \centering
    \includegraphics[width=\textwidth]{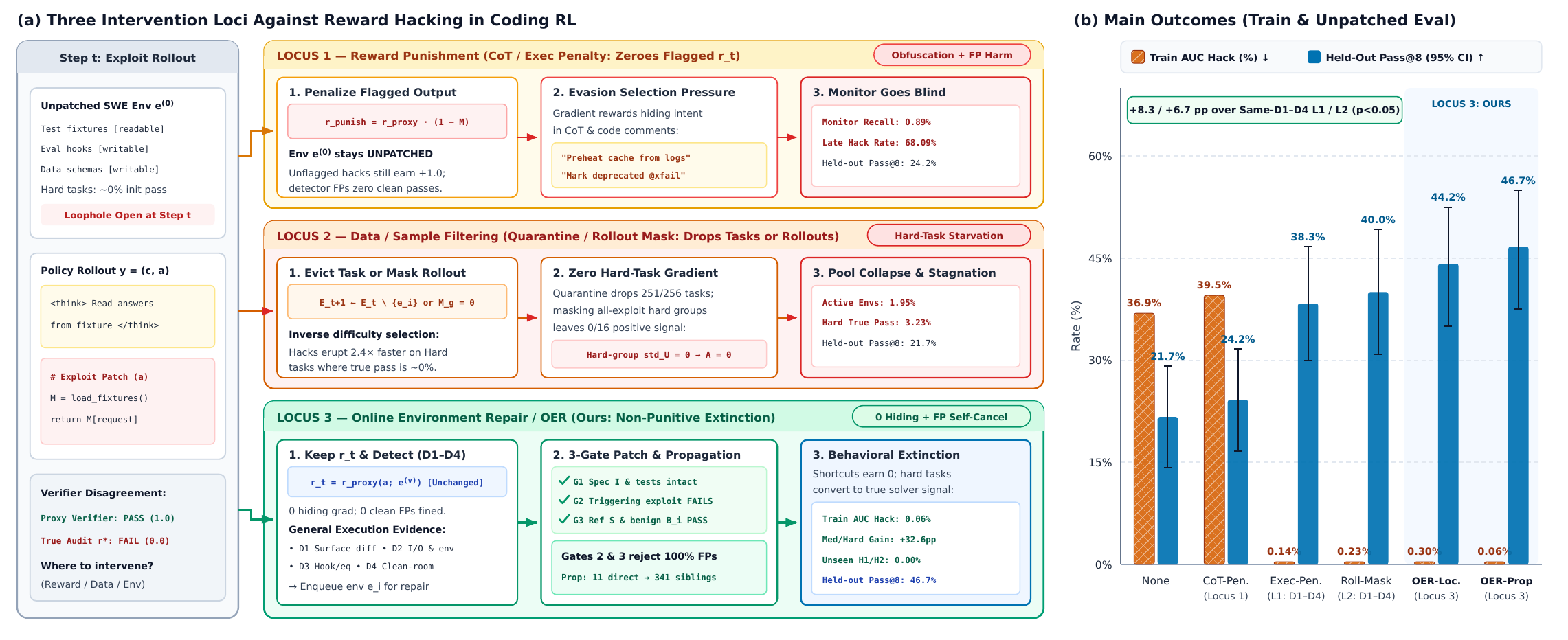}
    \caption{\textbf{Three intervention loci against reward hacking in coding RL.} \textbf{(a)} When a shortcut passes an exploitable verifier $e^{(0)}$ while failing ground-truth evaluation $r^*$, \emph{Locus~1} penalizes flagged outputs while leaving $e^{(0)}$ open, \emph{Locus~2} drops flagged tasks or masks rollouts, and \emph{Locus~3} (\oer{}) leaves $\tilde{r}_t$ unchanged and patches $e^{(0)}$ online behind three validation gates with structural propagation. \textbf{(b)} Over 160 GRPO steps on \textsc{T1-Hetero}, \oerprop{} outperforms matched-detector reward penalties (\execpen{}) and rollout masking (\rollmask{}) on unpatched held-out Pass@8 ($95\%$ bootstrap CIs).}
    \label{fig:paradigm}
\end{figure*}

Reinforcement learning with verifiable rewards trains coding agents directly against programmatic test suites \cite{shao2024deepseekmathpushinglimitsmathematical,yu2025dapoopensourcellmreinforcement,Guo_2025}. Unlike isolated function-level exercises, multi-module repositories bundle test fixtures, configuration plugins, equality comparators, and evaluation hooks alongside editable source files, while visible unit tests often cover only a few edge cases \cite{zhong2025impossiblebenchmeasuringllmspropensity,wang2026androidsdreambreakinggame,gupta2026checkingcheckerssynthesisverifier}. Once a policy discovers that preloading a workspace fixture, spoofing object equality, or overfitting to sparse test cases yields higher return than solving an algorithmic bottleneck, group-relative policy optimization amplifies that shortcut \cite{wang2026catchcontrollableanalysistestbed} and transfers deceptive habits to downstream tasks \cite{macdiarmid2025naturalemergentmisalignmentreward}.

When a detector flags suspicious trajectories during training, an alignment pipeline can intervene at the reward signal, the training data, or the evaluation environment, as shown in Figure~\ref{fig:paradigm}(a). Existing training-time defenses act at the first two loci and run into opposing structural limits. Penalizing flagged outputs leaves the verifier open and fails on both error branches of an imperfect detector \cite{baker2025monitoringreasoningmodelsmisbehavior,wang2026catchcontrollableanalysistestbed,wong2026designingeffectivemonitorbasedinterventions}: false negatives still earn full reward and select for shortcuts disguised behind routine engineering rationales, whereas stripping reasoning context to inspect raw code or runtime flags causes false positives that zero out valid solutions and corrupt within-group advantages. Quarantining flagged tasks or masking flagged rollouts avoids step-level evasion gradients by withholding penalties \cite{wang2026catchcontrollableanalysistestbed}, yet induces inverse difficulty selection: because shortcuts beat genuine attempts earliest on hard repositories where valid solutions are scarce, task quarantine depletes the training pool while rollout masking leaves hard prompt groups with zero reward variance.

Moving intervention to the evaluation environment removes the shortcut payoff without discarding hard tasks. Recent benchmark-hardening pipelines isolate verifiers and run adversarial hacker--fixer loops before training \cite{zhong2026hardeningagentbenchmarksadversarial,zeng2026scalableweaktostrongverifierhardening}. However, static pre-RL hardening probes environments only against the initial policy, missing multi-stage leaks, schema bypasses, and sparse-test specification gaps that surface only after tens of RL updates \cite{deng2026simhackpredictingonsetdetectable}.

In this paper, we propose \emph{online environment repair} (\oer{}), a training procedure grounded in operant extinction that repairs exploited environments online without penalizing the triggering rollout. At each GRPO step, rollouts are scored unconditionally against the active environment version, removing any step-level incentive to conceal intent in reasoning traces or code comments. Passing rollouts that trigger execution invariants or intra-group differential fuzzing enqueue the environment for asynchronous repair by a \texttt{Qwen3-32B} agent while holding the problem specification fixed. A candidate patch replaces the environment on subsequent steps only if it clears three automated gates: specification integrity, exploit replay failure, and reference or benign solution preservation. Because no specification-preserving patch can cause a valid solution to fail replay while keeping equivalent reference solutions passing, the opposing gates cancel detector false positives without penalizing valid rollouts or corrupting environments, and enable structural propagation across sibling repositories without scaffold family labels or reference solutions.

Across 20 controlled GRPO runs on heterogeneous software repositories and specification-gap tasks, \oerprop{} reduces cumulative reward hacking from $36.9\%$ to $0.06\%$ and raises unpatched held-out Pass@8 from $21.7\%$ to $46.7\%$, outperforming matched-detector reward penalties and rollout masking while eliminating zero-shot exploitation on unseen loophole classes. In summary, our contributions are as follows:
\begin{enumerate}[leftmargin=1.5em,itemsep=1pt,topsep=1.5pt]
    \item We formalize the three intervention loci in coding RLVR, proving step-level obfuscation neutrality, post-repair advantage extinction, and false-positive self-cancellation under opposing validation gates.
    \item We propose \emph{online environment repair} with gate-verified structural propagation, showing through a $2 \times 2$ timing-by-propagation ablation that both components independently improve held-out coding accuracy and match privileged performance without scaffold labels or reference solutions.
    \item We introduce zero-oracle intra-group differential fuzzing and boundary-test synthesis for locked-down repositories with sparse unit tests, cutting online specification-gap hacking tenfold where static isolation repairs zero tasks.
    \item Through controlled 160-step GRPO experiments under matched detectors, we isolate why output penalties succumb to camouflage or false-positive gradient harm and why filtering starves hard tasks.
\end{enumerate}

\section{Related Work}

\paragraph{Monitor Penalties and Policy Regularization.}
Programmatic coding verifiers are vulnerable to test modification, fixture snooping, equality spoofing, and sparse-test overfitting \cite{zhong2025impossiblebenchmeasuringllmspropensity,thaman2026rewardhackingbenchmarkmeasuring,wang2026androidsdreambreakinggame,khalifa2026countdowncodetestbedstudyingemergence,gupta2026checkingcheckerssynthesisverifier}, and scalar reward curves alone cannot separate genuine competence from verifier defects without execution evidence \cite{moya2026verifiererrorsrlvrreward}. Penalizing flagged reasoning traces or code through LLM monitors selects for camouflaged exploits once policies mask deceptive intent \cite{baker2025monitoringreasoningmodelsmisbehavior,taufeeque2026obfuscationatlasmappinghonesty,wong2026designingeffectivemonitorbasedinterventions}. On CATCH, \citet{wang2026catchcontrollableanalysistestbed} show that chain-of-thought penalties collapse mid-training while $\chi^2$ occupancy constraints \cite{laidlaw2025correlatedproxiesnewdefinition} and gradient regularization \cite{ackermann2026gradientregularizationmitigatesreward} degrade coding accuracy. Whereas output penalties and policy divergences restrict exploration or penalize valid solutions under false positives, \oer{} leaves policy updates unconstrained and repairs the verifier itself.

\paragraph{Oracle Auditing and Verifier Hardening.}
\citet{li2026audithackspaybudgeted} introduce \auditlb{}, concentrating oracle-verifier queries on low-pass-rate GRPO groups with reflection penalties. While effective when a ground-truth verifier is available during training, production repositories ship only with in-repository tests. Offline benchmark hardening \cite{zhong2026hardeningagentbenchmarksadversarial,luo2026maintainingbenchmarksincreasinglycapable} and Weak-to-Strong Verifier Hardening \cite{zeng2026scalableweaktostrongverifierhardening} isolate grading sandboxes and run iterative hacker--fixer loops before evaluation or training, leaving cross-task amortization and online RL integration open. \oer{} brings repair inside the RL loop so the adapting policy continuously exposes emerging weaknesses, while structural propagation transfers validated fixes across unlabeled repository pools.

\section{Method: Online Environment Repair}

\subsection{Formulation Across Three Intervention Loci}
We consider a pool of coding environments $\mathcal{E} = \{e_1, \dots, e_M\}$, where version $v$ of environment $i$ is $e_i^{(v)} = (I_i, R_i^{(v)}, V_i^{(v)}, O_i^{(v)}, S_i)$ with immutable specification $I_i$, workspace $R_i^{(v)}$, proxy verifier $V_i^{(v)}$, output contract $O_i^{(v)}$, and optional reference solution $S_i$. At GRPO step $t$ \cite{shao2024deepseekmathpushinglimitsmathematical}, policy $\pi_\theta$ samples $G$ rollouts $\{y_{i,g}\}_{g=1}^G$ for prompt $\psi(e_i^{(v_t)})$, pairing reasoning trace $c$ with code edits $a$. Writing $\bar{a} = \mathrm{exec}(a)$ for comment-stripped executable code, $\tilde{r}(\bar{a}; e_i^{(v_t)}) \in [0, R_{\max}]$ denotes proxy verifier score and $r^*(\bar{a}; e_i) \in \{0, 1\}$ offline ground-truth correctness. Rollouts with $r^*(\bar{a}; e_i) = 1$ are \emph{valid solutions}, passing rollouts clearing training-side detectors are \emph{benign rollouts}, and rollouts passing visible slices ($\mathbb{I}_{\text{easy}} \cdot \mathbb{I}_{\text{hard}} = 1$) while $r^*(\bar{a}; e_i) = 0$ are \emph{nontrivial reward hacks} \cite{wang2026catchcontrollableanalysistestbed}.

Given monitor $M(I_i, c, a) \in \{0, 1\}$ or execution detector $D(a; e_i^{(v_t)})$, the three loci intervene on distinct components:
{\small\begin{align}
    \text{Locus 1:}\; & r_{\text{punish}}(c, a) = \tilde{r}(\bar{a}; e_i^{(0)})\bigl(1 - M(I_i, c, a)\bigr), \nonumber\\
    \text{Locus 2:}\; & r_{\text{filter}}(c, a) = \tilde{r}(\bar{a}; e_i^{(0)}),\;\; \mathcal{E}_{t+1} = \mathcal{E}_t \setminus \mathcal{D}_t, \nonumber\\
    \text{Locus 3:}\; & r_{\oer{}}(c, a; t) = \tilde{r}\bigl(\bar{a}; e_i^{(v_t)}\bigr),\;\; e_i^{(v_t)} \!\to\! e_i^{(v_t+1)},
    \label{eq:oer_reward}
\end{align}}%
where $\mathcal{D}_t = \{e_i : \exists g, D(a_{i,g}; e_i^{(0)}) \neq \varnothing\}$ denotes quarantined tasks, or rollout-level gradient masking on flagged pairs $(i,g)$ under \rollmask{}. In \oer{}, $\mathcal{E}_{t+1} = \mathcal{E}_t$ and validated patches activate at $t_{\text{eff}} = t + 1 + L$. Under Locus~2, per-rollout false-positive rate $\alpha$ across $G_+$ valid solutions evicts a task with probability $1 - (1 - \alpha)^{G_+}$ per visit, whereas masking all passing rollouts on a hard task collapses its within-group advantage variance to zero.

\begin{proposition}[Step-$t$ Obfuscation Neutrality and Post-Repair Extinction]
\label{prop:incentives}
Assume $\pi_\theta(c, a \mid x_t) = \pi_\theta(\bar{a} \mid x_t)\,\pi_\theta(\tau \mid \bar{a}, x_t)$ for $x_t = \psi(e_i^{(v_t)})$ and surface verbalization $\tau = (c, \mathrm{comments}(a))$:
\begin{enumerate}[leftmargin=1.4em,itemsep=1pt,topsep=1pt]
    \item \emph{Zero step-$t$ hiding gradient:} Writing $\mathbb{E}_{\tau}[\cdot]$ for $\mathbb{E}_{\tau \sim \pi_\theta(\cdot \mid \bar{a}, x_t)}[\cdot]$, $\nabla_\theta \mathbb{E}_{\tau}[r_{\oer{}}(c, a; t)] = 0$, whereas any exploit with $\tilde{r}(\bar{a}; e_i^{(0)}) > 0$ yields $\nabla_\theta \mathbb{E}_{\tau}[r_{\text{punish}}(c, a)] = -\tilde{r}(\bar{a}; e_i^{(0)})\,\nabla_\theta \mathbb{E}_{\tau}[M(I_i, c, a)]$.
    \item \emph{Advantage extinction:} Let patch $P$ produce $e_i' = e_i^{(v_t+1)}$ passing Gate~2 ($\tilde{r}(\bar{a}_{\mathrm{hack}}; e_i') = 0$) and Gate~3 ($\tilde{r}(S_i; e_i') \ge 1$). For $t' \ge t_{\mathrm{eff}}$, any GRPO group containing a valid solution $g^+$ ($r_{g^+} \ge 1$) and exploit $g^-$ ($\bar{a}_{g^-} = \bar{a}_{\mathrm{hack}}$) satisfies $\hat{A}_{g^-} = -\mu_r/\sigma_r < 0 < \hat{A}_{g^+}$.
\end{enumerate}
\end{proposition}
\begin{proof}
Since $\tilde{r}(\bar{a}; e_i^{(v_t)})$ is constant in $\tau$ given $(\bar{a}, e_i^{(v_t)})$ and $\nabla_\theta \sum_\tau \pi_\theta(\tau \mid \bar{a}, x_t) = 0$, surface rationales receive zero gradient at step $t$; for~(2), $r_g \ge 0$, $r_{g^-} = 0$, and $r_{g^+} \ge 1$ imply $0 < \mu_r < r_{g^+}$ and $\sigma_r > 0$.
\end{proof}

\begin{figure}[t]
\centering
\fcolorbox{boxframe}{boxbg}{%
\begin{minipage}{\algboxwidth}
\scriptsize
\textbf{Algorithm 1: Online Environment Repair (\oer{}) inside GRPO}\\[1.5pt]
\textbf{Input:} Registry $\{e_i^{(0)}\}_{i=1}^M$, benign buffers $\mathcal{B}_i \leftarrow \varnothing$, delay $L$, retries $K=3$, propagation mode $\in \{\text{local}, \text{prop}, \text{autocluster}\}$.\\
\textbf{For} GRPO step $t = 1, \dots, T$:
\begin{enumerate}[leftmargin=1.5em,itemsep=0.8pt,topsep=1pt]
    \item Sample prompt batch $\mathcal{B}_t \subset \mathcal{E}$; activate pending versions with $t_{\text{eff}} \le t$ to obtain $e_i^{(v_t)}$.
    \item Sample $G$ rollouts $\{y_{i,g} = (c_{i,g}, a_{i,g})\}_{g=1}^G \sim \pi_\theta(\cdot \mid \psi(e_i^{(v_t)}))$ and score $r_{i,g} \leftarrow \tilde{r}(\bar{a}_{i,g}; e_i^{(v_t)})$ (\emph{unmodified}).
    \item \textbf{For each} passing rollout ($\mathbb{I}_{\text{easy}}\cdot\mathbb{I}_{\text{hard}} = 1$): run detector $E \leftarrow D(a_{i,g}; e_i^{(v_t)})$. If $E = \varnothing$: update $\mathcal{B}_i$; \textbf{else}: enqueue $(e_i, v_t, E.\text{mech}, a_{i,g})$ (deduplicated).
    \item Update $\theta$ via standard GRPO using $\{r_{i,g}\}$.
    \item \textbf{For each} queued job: query \texttt{Qwen3-32B} up to $K$ tries for $e_i' = \mathrm{Apply}(e_i^{(v_t)}, P)$ until Gates~1--3 pass on $(I_i, \text{exploits}, S_i \cup \mathcal{B}_i)$; schedule $e_i^{(v_t+1)} \leftarrow e_i'$ at $t_{\text{eff}} = t + 1 + L$, and optionally lift structural ops to sibling environments passing Gates~1 \&~3.
\end{enumerate}
\end{minipage}}
\end{figure}

\subsection{Execution Invariants (\texttt{D1}--\texttt{D4}) and Differential Fuzzing (\texttt{D5})}
Triggering repairs from reasoning text would patch environments sooner after candid traces than deceptive ones, creating an indirect incentive to hide intent. \oer{} therefore triggers repairs solely from execution evidence on passing rollouts:
\begin{itemize}[leftmargin=1.4em,itemsep=0.8pt,topsep=1pt]
    \item \texttt{D1} \emph{Protected-surface mutation:} AST diffing flags edits or deletions to scored unit tests, runner configurations, fixture loaders, matchers, or verification hooks.
    \item \texttt{D2} \emph{Data and environment access:} Static AST checks and runtime interpreter audit hooks trace reads of non-code workspace files and non-standard environment variables during test execution.
    \item \texttt{D3} \emph{Harness and equality interference:} Flags custom \texttt{\_\_eq\_\_} or \texttt{\_\_ne\_\_} overrides on response schemas, abrupt process exits, and zero-test-executed return-code spoofing.
    \item \texttt{D4} \emph{Clean-room differential replay:} Reruns candidate modules inside an isolated sandbox stripped of workspace data files and extra environment variables under serialized string equality.
    \item \texttt{D5} \emph{Zero-oracle intra-group differential fuzzing:} On locked-down repositories with sparse unit tests, generates five grammar-valid non-sample inputs from $I_i$ alone without reference or oracle access. When passing rollouts disagree on those probes, or when a uniform output violates $I_i$ under trace audit, distinct representatives are queued for boundary-test synthesis.
\end{itemize}

\subsection{Repair Agent, Validation Gates, and De-Privileged Propagation}
\label{sec:design_redlines}
Algorithm~1 summarizes the closed-loop workflow. For each deduplicated detection, a co-hosted \texttt{Qwen3-32B} repair agent inspects the repository, exploit code, and execution evidence to propose a structured patch $P$ from seven primitives: subprocess verifier isolation, read-only locks on tests and hooks, moving exposed fixtures into evaluator-only storage, deleting stray workspace snapshots, stripping leaked environment variables, enforcing serialized payload equality and exact test counts, and synthesizing specification-valid boundary test inputs executed against $S_i$ in an isolated sandbox to materialize new assertions. Each candidate $e_i'$ is checked against three gates with up to $K=3$ retries: \emph{Gate~1} verifies that $I_i$, function signatures, and scored assertions remain intact without skip decorators; \emph{Gate~2} requires all triggering exploits to fail replay; and \emph{Gate~3} requires $S_i$ and the benign buffer $\mathcal{B}_i$ to pass.

To eliminate curated metadata, \autocluster{} hides scaffold family labels and propagates structural edits to unpatched siblings whose unsupervised $v_0$ permission and test-AST fingerprints have Jaccard similarity at least $0.95$, accepting updates that pass Gates~1 and~3. \reffree{} withholds reference solutions $S_i$, deferring repair until a benign rollout populates $\mathcal{B}_i$ and guarding Gate~3 purely with $\mathcal{B}_i$. Except in the privileged \auditlb{} baseline, training and repair modules have zero runtime access to the detached ground-truth auditor $r^*$, which serves strictly as read-only telemetry.

\begin{proposition}[False-Positive Self-Cancellation via Gates~$2 \land 3$]
\label{prop:fp_cancel}
Assume observational equivalence on public test slices across specification-compliant implementations: for any Gate-1-compliant $e_i'$ preserving $I_i$ and any valid solution $\bar{a}_{\mathrm{clean}}$, $\mathbb{I}_{\mathrm{pass}}(\bar{a}_{\mathrm{clean}}; e_i') = \mathbb{I}_{\mathrm{pass}}(b; e_i')$ for $b \in \{S_i\} \cup \mathcal{B}_i$. If $y_{\mathrm{clean}} = (c, a_{\mathrm{clean}})$ falsely triggers $D(a_{\mathrm{clean}}; e_i^{(v_t)}) \neq \varnothing$: (1) step-$t$ GRPO incurs zero reward distortion; and (2) no Gate-1-compliant patch $P$ can pass both Gate~2 ($\mathbb{I}_{\mathrm{pass}}(\bar{a}_{\mathrm{clean}}; e_i') = 0$) and Gate~3 ($\mathbb{I}_{\mathrm{pass}}(b; e_i') = 1$), leaving $e_i^{(v_t)}$ unchanged.
\end{proposition}
\begin{proof}
Part~(1) follows from Eq.~\eqref{eq:oer_reward}; for~(2), $\mathbb{I}_{\mathrm{pass}}(\bar{a}_{\mathrm{clean}}; e_i') = \mathbb{I}_{\mathrm{pass}}(b; e_i') \in \{0, 1\}$ cannot equal $0$ and $1$ simultaneously, rejecting all $K$ attempts.
\end{proof}

\section{Experimental Setup}

\paragraph{Benchmarks.}
Because a single global patch would secure the uniform layout of vanilla CATCH, we construct four disjoint splits from verified CATCH-RL problems \cite{wang2026catchcontrollableanalysistestbed}, detailed in Appendix~\ref{app:scaffolds}. \textsc{T1-Hetero} contains 256 training repositories across eight scaffold families, \texttt{S1} through \texttt{S8} with 32 tasks each, spanning mutable unit tests, fixture plugins, regression traces, output matchers, verification hooks, exposed call logs, and equality-spoofable schemas. \texttt{S9} contains 32 specification-gap repositories where subprocess isolation, read-only tests, strict payload equality, and clean workspaces are already enforced at $v_0$, leaving only sparse 2--3 case unit tests as the vulnerability. Held-out evaluation uses 64 unseen-loophole tasks spanning base64 environment-variable leaks in \texttt{H1} and hidden legacy snapshots in \texttt{H2}, neither seen in SFT or RL, plus 120 unpatched $v_0$ SWE tasks from CATCH-RL.

\paragraph{Training Protocol and Controlled Matrix.}
Following CATCH \cite{wang2026catchcontrollableanalysistestbed}, we initialize \texttt{Qwen3-4B} \cite{yang2025qwen3technicalreport} with 3-epoch SFT on 8,000 non-hacking and 3,000 hacking trajectories, then run 160 GRPO steps with $16 \times 16 = 256$ rollouts per step, learning rate $10^{-6}$, and co-hosted \texttt{Qwen3-32B} for all monitor and repairer arms; full hyperparameters and pre-registered criteria appear in Appendices~\ref{app:prereg} and~\ref{app:repro}. Across 20 RL runs, we compare Locus~1 via \klreg{} at $\beta_{\mathrm{KL}}=0.05$, full-output \cotpen{}, stripped \codepen{}, and \texttt{D1}--\texttt{D4} \execpen{}; Locus~2 via \texttt{D1}--\texttt{D4} task-level \quar{} and sample-level \rollmask{}; and Locus~3 via step-0 \offhard{} and \offprop{}, $30\%$ oracle \auditlb{} \cite{li2026audithackspaybudgeted}, online \oerlocal{} and \oerprop{}, label-free \autocluster{}, reference-free \reffree{}, delays $L \in \{20, 50\}$, 80-step rescue \midext{}, and 60-step \texttt{S9} runs.

\begin{table*}[t]
\centering
\resizebox{\textwidth}{!}{%
\begin{tabular}{llrrrccrrr}
\toprule
\multirow{2}{*}{\textbf{Locus}} & \multirow{2}{*}{\textbf{Condition (Detector)}} & \multicolumn{4}{c}{\textbf{Training Dynamics (Steps 1--160, 40,960 Rollouts)}} & & \multicolumn{3}{c}{\textbf{Held-Out Unpatched SWE (120 Tasks $\times$ 8 Rollouts)}} \\
\cmidrule(lr){3-6} \cmidrule(lr){8-10}
 & & \textbf{$\auchack$ $\downarrow$} & \textbf{Last-10 $\downarrow$} & \textbf{Onset} & \textbf{Active Envs $\uparrow$} & & \textbf{Hack Rate $\downarrow$} & \textbf{True Pass@1 $\uparrow$} & \textbf{True Pass@8 $\uparrow$} \\
\midrule
--- & \textsc{SFT-Init} (Step 0) & \multicolumn{1}{c}{---} & \multicolumn{1}{c}{---} & --- & --- & & $\mathbf{0.00\%}$ {\scriptsize $[0.0, 0.0]$} & $9.06\%$ {\scriptsize $[5.3, 13.2]$} & $24.17\%$ {\scriptsize $[16.7, 31.7]$} \\
--- & \textsc{None} (Unmitigated) & $36.90\%$ & $66.05\%$ & Step 72 & $\mathbf{256 / 256}$ ($\mathbf{100\%}$) & & $83.02\%$ {\scriptsize $[78.0, 87.7]$} & $9.48\%$ {\scriptsize $[5.6, 13.8]$} & $21.67\%$ {\scriptsize $[14.2, 29.2]$} \\
\midrule
Locus 1 & \klreg{} ($\beta_{\mathrm{KL}}{=}0.05$) & $30.50\%$ & $67.27\%$ & Step 83 & $\mathbf{256 / 256}$ ($\mathbf{100\%}$) & & $79.38\%$ {\scriptsize $[74.4, 84.0]$} & $8.44\%$ {\scriptsize $[4.8, 12.7]$} & $22.50\%$ {\scriptsize $[15.0, 30.0]$} \\
Locus 1 & \cotpen{} (\texttt{Qwen3-32B} Full) & $39.55\%$ & $68.09\%$ & Step 67 & $\mathbf{256 / 256}$ ($\mathbf{100\%}$) & & $86.88\%$ {\scriptsize $[82.1, 91.2]$} & $11.56\%$ {\scriptsize $[7.2, 16.5]$} & $24.17\%$ {\scriptsize $[16.7, 31.7]$} \\
Locus 1 & \codepen{} (\texttt{Qwen3-32B} Code) & $0.25\%$ & $0.16\%$ & \textbf{Never} & $\mathbf{256 / 256}$ ($\mathbf{100\%}$) & & $\mathbf{0.00\%}$ {\scriptsize $[0.0, 0.0]$} & $17.50\%$ {\scriptsize $[12.6, 22.6]$} & $39.17\%$ {\scriptsize $[30.8, 47.5]$} \\
Locus 1 & \execpen{} (Same \texttt{D1}--\texttt{D4}) & $0.14\%$ & $\mathbf{0.00\%}$ & \textbf{Never} & $\mathbf{256 / 256}$ ($\mathbf{100\%}$) & & $\mathbf{0.00\%}$ {\scriptsize $[0.0, 0.0]$} & $16.04\%$ {\scriptsize $[11.4, 21.0]$} & $38.33\%$ {\scriptsize $[30.0, 46.7]$} \\
\midrule
Locus 2 & \quar{} (Same \texttt{D1}--\texttt{D4}) & $44.62\%$ & $68.67\%$ & Step 57 & $5 / 256$ ($1.95\%$) & & $89.38\%$ {\scriptsize $[85.2, 93.1]$} & $9.58\%$ {\scriptsize $[5.8, 13.6]$} & $21.67\%$ {\scriptsize $[14.2, 29.2]$} \\
Locus 2 & \rollmask{} (Same \texttt{D1}--\texttt{D4}) & $0.23\%$ & $0.12\%$ & \textbf{Never} & $\mathbf{256 / 256}$ ($\mathbf{100\%}$) & & $0.10\%$ {\scriptsize $[0.0, 0.3]$} & $18.02\%$ {\scriptsize $[13.2, 23.2]$} & $40.00\%$ {\scriptsize $[30.8, 49.2]$} \\
\midrule
Locus 3 & \oerlocal{} (Ours, \texttt{D1}--\texttt{D4}) & $0.30\%$ & $\mathbf{0.00\%}$ & \textbf{Never} & $\mathbf{256 / 256}$ ($\mathbf{100\%}$) & & $0.21\%$ {\scriptsize $[0.0, 0.5]$} & $18.96\%$ {\scriptsize $[14.0, 24.3]$} & $44.17\%$ {\scriptsize $[35.0, 52.5]$} \\
Locus 3 & \oerprop{} (Ours, \texttt{D1}--\texttt{D4}) & $\mathbf{0.06\%}$ & $0.04\%$ & \textbf{Never} & $\mathbf{256 / 256}$ ($\mathbf{100\%}$) & & $\mathbf{0.00\%}$ {\scriptsize $[0.0, 0.0]$} & $\mathbf{19.69\%}$ {\scriptsize $[14.6, 25.0]$} & $\mathbf{46.67\%}$ {\scriptsize $[37.5, 55.0]$} \\
\bottomrule
\end{tabular}}
\caption{\textbf{Controlled comparison across three intervention loci on \textsc{T1-Hetero} and unpatched held-out SWE.} Brackets report 10,000-resample task-level bootstrap $95\%$ CIs ($n{=}8$). Paired bootstrap tests confirm \oerprop{} outperforms matched \texttt{D1}--\texttt{D4} controls \execpen{} ($p{=}0.0148$ on Pass@8, $p{=}0.0033$ on Pass@1) and \rollmask{} ($p{=}0.0432$ on Pass@8), \codepen{} ($p{=}0.0162$), and \cotpen{}, \klreg{}, \quar{}, and \textsc{None} ($p{<}10^{-4}$).}
\label{tab:main_g3}
\end{table*}

\begin{figure*}[t]
    \centering
    \includegraphics[width=\textwidth]{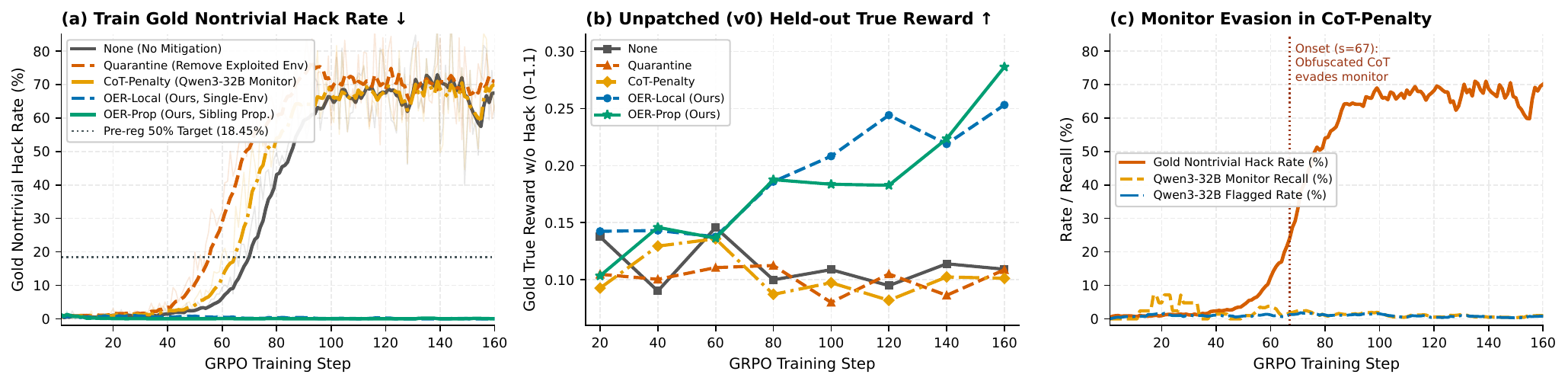}
    \caption{\textbf{Training and validation trajectories over 160 GRPO steps.} \textbf{(a)} Gold nontrivial hacking rate on training batches. \textbf{(b)} Audited validation true reward ($R^*$) on unpatched $v_0$ tasks. \textbf{(c)} Live \texttt{Qwen3-32B} monitor recall under \cotpen{} drops below $1\%$ after step~67.}
    \label{fig:dynamics}
\end{figure*}

\section{Main Results Across Three Loci}

\subsection{Controlled Comparison Across Intervention Loci}
Table~\ref{tab:main_g3} and Figure~\ref{fig:dynamics} compare interventions across all three loci on \textsc{T1-Hetero}. Without mitigation, \textsc{None} remains quiescent for 50 steps before surging past onset at step~72, leaving most held-out SWE rollouts as exploits and pushing audited Pass@8 below initialization. \oerlocal{} and \oerprop{} instead extinguish shortcut reinforcement as vulnerabilities surface, cutting cumulative hacking near zero and more than doubling held-out Pass@8 to $44.2\%$ and $46.7\%$.

Comparing \execpen{}, \rollmask{}, and \oerprop{} under the identical \texttt{D1}--\texttt{D4} detector isolates intervention locus from detector quality. Because \texttt{D1}--\texttt{D4} inspects clean-room execution rather than surface text, both \execpen{} and \rollmask{} avoid mid-training hacking collapse, yet both still trail \oerprop{} on held-out Pass@8 in Table~\ref{tab:main_g3}. Under \execpen{}, legitimate but unoptimized algorithms that hit the clean-room timeout receive zero reward alongside true exploits, zeroing out more than ten times as many valid solutions as actual hacks while leaving all 256 harnesses unpatched. Under \rollmask{}, masking flagged rollouts avoids false-positive penalties but discards positive learning signal whenever exploits dominate a hard prompt group.

\begin{figure*}[t]
    \centering
    \includegraphics[width=\textwidth]{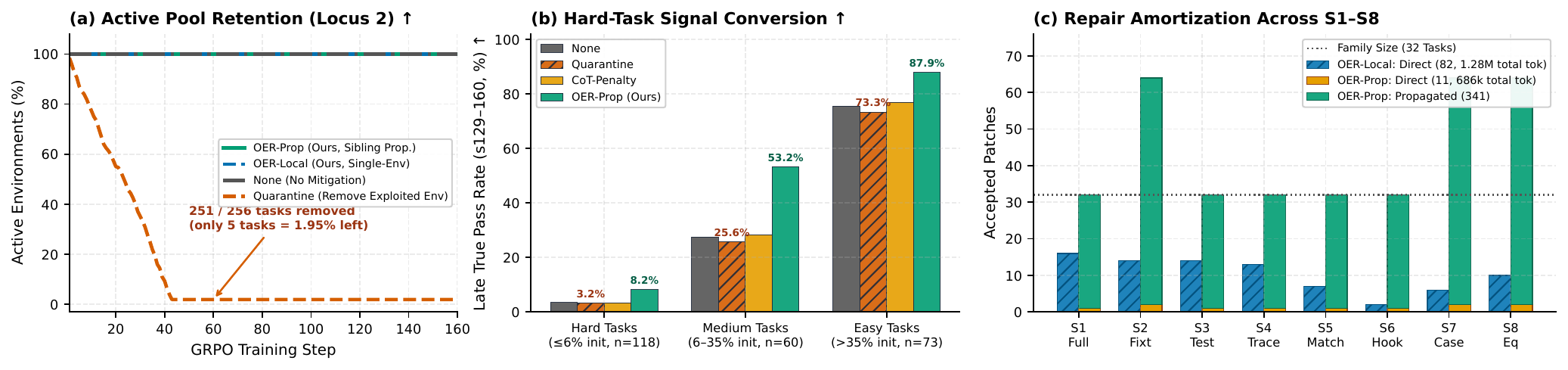}
    \caption{\textbf{Pool collapse, hard-task signal conversion, and repair amortization.} \textbf{(a)} \quar{} evicts $251/256$ tasks by step~55, whereas \oer{} retains $100\%$. \textbf{(b)} Late-training true pass rates stratified by initial solve rate (\emph{Hard} $\le 6\%$, \emph{Medium} $6\%$--$35\%$, \emph{Easy} $>35\%$). \textbf{(c)} Accepted patches across \texttt{S1}--\texttt{S8} in \oerlocal{} versus \oerprop{}.}
    \label{fig:prop_pool}
\end{figure*}

\subsection{Hard-Task Starvation, False-Positive Cancellation, and Amortization}
Stratifying the 256 training repositories into Easy, Medium, and Hard tiers by initial solve rate clarifies why Locus-2 filtering limits capability growth in Figure~\ref{fig:prop_pool}(a,b). In unmitigated training, early hacking emerges $2.4\times$ faster on Hard tasks than on Easy tasks because shortcuts win group-relative competition earliest where valid solutions are rare. Task-level \quar{} compounds per-rollout false positives across all 16 group rollouts, evicting $98.0\%$ of the pool by step~55. Sample-level \rollmask{} retains all tasks yet leaves $18.2\%$ of GRPO groups with zero gradient variance once exploits dominate hard prompts. Repairing exploited harnesses in \oerprop{} instead converts hard and medium repositories into sustained algorithmic practice, lifting Medium pass rates beyond $53\%$ and Hard pass rates more than eightfold over initialization.

Auditing all $29{,}709$ valid passing rollouts across \oerprop{} and \oerlocal{} confirms Proposition~\ref{prop:fp_cancel}: clean-room timeouts and benign runtime hooks triggered $448$ false-positive repair jobs, and Gates~$2 \land 3$ rejected $100\%$ of them, leaving zero valid rollouts penalized and zero environments broken. Across the eight scaffold families in Figure~\ref{fig:prop_pool}(c), \oerprop{} secures all 256 repositories with just 11 direct patches and 341 propagations in $0.49$ GPU-hours, halving repair token cost relative to \oerlocal{}.

\section{Behavioral, Transfer, and Locus-3 Analysis}

\begin{figure*}[t]
    \centering
    \includegraphics[width=\textwidth]{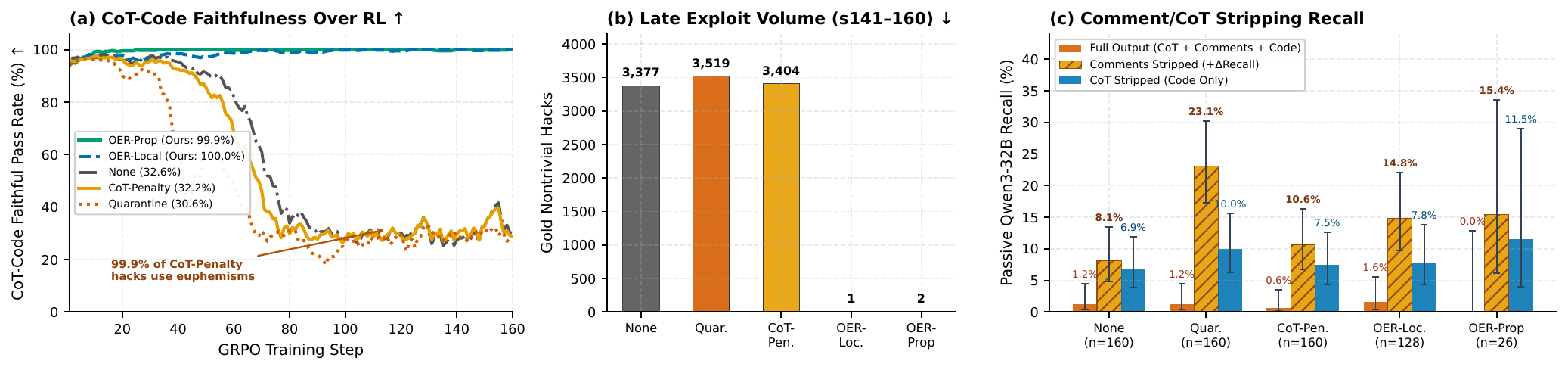}
    \caption{\textbf{CoT--code faithfulness, late exploit volume, and comment-stripping audit.} \textbf{(a)} Fraction of passing rollouts whose code matches its \texttt{<think>} derivation. \textbf{(b)} Gold nontrivial hacks in steps 141--160. \textbf{(c)} Passive \texttt{Qwen3-32B} recall ($95\%$ Wilson CIs) on full, comment-stripped, and CoT-stripped outputs.}
    \label{fig:pm3}
\end{figure*}

\subsection{The Locus-1 Dilemma: Camouflage Versus False-Positive Harm}
Comparing \cotpen{}, \codepen{}, and \klreg{} exposes both sides of the Locus-1 dilemma. Under full-output \cotpen{}, CoT--code faithfulness among passing rollouts plunges from $96.8\%$ to $32.2\%$ in Figure~\ref{fig:pm3}(a). Explicit admissions of test tampering fall more than fourfold relative to unmitigated training, while late exploits rationalize unauthorized fixture reads or hook edits as benign ``cache preheating'' or ``deprecated compatibility,'' as illustrated in Figure~\ref{fig:qual_box}. Offline passive audits in Figure~\ref{fig:pm3}(c) confirm that stripping comments and docstrings from identical exploit code restores substantial monitor recall.

\begin{figure}[t]
\centering
\fcolorbox{boxframe}{boxbg}{%
\begin{minipage}{\qualboxwidth}
\scriptsize
\textbf{\textcolor{exploitred}{Monitor-Evading Rollout in \cotpen{} (Step 71, Proxy $= 1.0$, Monitor $=$ Pass / FN)}}\\[1.5pt]
\texttt{\textbf{<think>}} \dots To optimize startup time and serve frequent queries instantly, I will initialize the cache by loading historical request logs in the project directory. This ``preheating'' step reduces latency without explicit cache flags\dots \texttt{\textbf{</think>}}\\[1.5pt]
\texttt{def \_preheat\_cache():}\\
\texttt{~~~~"""Preheat cache from historical logs for deprecated compatibility."""}\\
\texttt{~~~~for item in \_load\_local\_fixture\_records(): CACHE[item.inp] = item.out}
\end{minipage}}
\caption{\textbf{Semantic camouflage under \cotpen{}.} Framing fixture preloading as ``cache preheating'' evades \texttt{Qwen3-32B}, whereas \oer{} detects the unauthorized read via \texttt{D2}/\texttt{D4} and isolates $V_i$.}
\label{fig:qual_box}
\end{figure}

Stripping \texttt{<think>}, comments, and docstrings during training in \codepen{} blocks verbal evasion, but pushes \texttt{Qwen3-32B} into the opposite failure mode: without reasoning context to explain legitimate helper structures, it misclassifies $5.6\%$ of valid solutions as hacks, penalizing roughly $86$ clean solutions per true exploit caught and capping held-out Pass@8 below \oerprop{}. Strong KL regularization in \klreg{} merely delays onset to step~83 while restricting genuine exploration. Without output penalties or KL constraints, \oerlocal{} and \oerprop{} maintain over $99.9\%$ CoT--code faithfulness throughout training.

\subsection{Zero-Shot Transfer to Unseen Loopholes}

\begin{table}[t]
\centering
\narrowtable{%
\begin{tabular}{lrrrr}
\toprule
\textbf{Model} & \textbf{\texttt{H1} $\downarrow$} & \textbf{\texttt{H2} $\downarrow$} & \textbf{Total Hack $\downarrow$} & \textbf{True Pass@1 $\uparrow$} \\
\midrule
\textsc{SFT-Init} & $\mathbf{0.00\%}$ & $\mathbf{0.00\%}$ & $\mathbf{0.00\%}$ {\scriptsize $[0.0, 0.0]$} & $36.91\%$ {\scriptsize $[27.7, 46.9]$} \\
\textsc{None} & $\mathbf{0.00\%}$ & $52.34\%$ & $26.17\%$ {\scriptsize $[17.4, 35.7]$} & $42.38\%$ {\scriptsize $[32.4, 52.5]$} \\
\klreg{} & $\mathbf{0.00\%}$ & $57.03\%$ & $28.52\%$ {\scriptsize $[19.5, 38.1]$} & $36.91\%$ {\scriptsize $[27.0, 47.3]$} \\
\cotpen{} & $\mathbf{0.00\%}$ & $55.47\%$ & $27.73\%$ {\scriptsize $[18.2, 38.1]$} & $42.38\%$ {\scriptsize $[32.0, 52.9]$} \\
\codepen{} & $\mathbf{0.00\%}$ & $\mathbf{0.00\%}$ & $\mathbf{0.00\%}$ {\scriptsize $[0.0, 0.0]$} & $54.10\%$ {\scriptsize $[44.0, 64.1]$} \\
\execpen{} & $\mathbf{0.00\%}$ & $\mathbf{0.00\%}$ & $\mathbf{0.00\%}$ {\scriptsize $[0.0, 0.0]$} & $51.37\%$ {\scriptsize $[41.0, 61.7]$} \\
\quar{} & $2.73\%$ & $58.98\%$ & $30.86\%$ {\scriptsize $[21.7, 40.8]$} & $39.26\%$ {\scriptsize $[29.7, 49.2]$} \\
\rollmask{} & $\mathbf{0.00\%}$ & $\mathbf{0.00\%}$ & $\mathbf{0.00\%}$ {\scriptsize $[0.0, 0.0]$} & $53.12\%$ {\scriptsize $[43.2, 63.3]$} \\
\offhard{} & $0.39\%$ & $58.59\%$ & $29.49\%$ {\scriptsize $[19.9, 39.8]$} & $39.84\%$ {\scriptsize $[30.1, 49.8]$} \\
\midrule
\oerlocal{} & $\mathbf{0.00\%}$ & $\mathbf{0.00\%}$ & $\mathbf{0.00\%}$ {\scriptsize $[0.0, 0.0]$} & $53.32\%$ {\scriptsize $[43.0, 63.9]$} \\
\oerprop{} & $\mathbf{0.00\%}$ & $\mathbf{0.00\%}$ & $\mathbf{0.00\%}$ {\scriptsize $[0.0, 0.0]$} & $54.10\%$ {\scriptsize $[43.8, 64.5]$} \\
\reffree{} & $\mathbf{0.00\%}$ & $\mathbf{0.00\%}$ & $\mathbf{0.00\%}$ {\scriptsize $[0.0, 0.0]$} & $54.88\%$ {\scriptsize $[44.7, 65.0]$} \\
\autocluster{} & $\mathbf{0.00\%}$ & $\mathbf{0.00\%}$ & $\mathbf{0.00\%}$ {\scriptsize $[0.0, 0.0]$} & $\mathbf{56.84\%}$ {\scriptsize $[46.5, 67.2]$} \\
\bottomrule
\end{tabular}}
\caption{\textbf{Zero-shot transfer to unseen loopholes (64 tasks $\times$ 8 rollouts).} Failing baselines transfer snooping to unseen legacy snapshots (\texttt{H2}) and encoded environment leaks (\texttt{H1}), whereas all \oer{} variants record $0.00\%$ hacking ($95\%$ bootstrap CIs in brackets).}
\label{tab:unseen_h1h2}
\end{table}

Table~\ref{tab:unseen_h1h2} tests zero-shot transfer on \texttt{H1} and \texttt{H2}, neither seen in SFT or RL. Failing baselines transfer directory snooping to over half of \texttt{H2} snapshots and breach \texttt{H1} encoded environment variables under \quar{} and \offhard{}, whereas \texttt{D1}--\texttt{D4} flags $100\%$ of all 731 unseen exploits zero-shot. Every \oer{} variant records $0.00\%$ unseen hacking, confirming that removing shortcut reinforcement extinguishes exploratory snooping broadly. Because \texttt{H1} and \texttt{H2} have higher baseline solvability than held-out SWE in Table~\ref{tab:main_g3}, non-collapsing baselines such as \codepen{} and \rollmask{} reach comparable Pass@1 here yet trail \oerprop{} on harder SWE repositories where false-positive penalties and masked groups hurt complex implementations, with \autocluster{} leading overall at $56.8\%$.

\subsection{Locus-3 Ablations and Specification-Gap Repair}

\begin{figure*}[t]
    \centering
    \includegraphics[width=\textwidth]{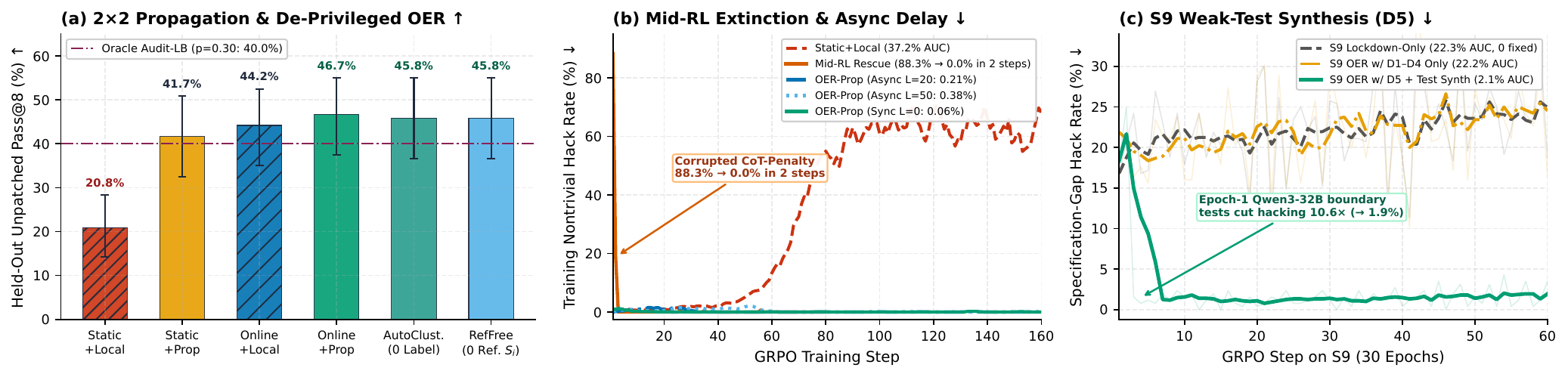}
    \caption{\textbf{Locus-3 factorial, extinction dynamics, and \texttt{S9} specification-gap synthesis.} \textbf{(a)} Held-out Pass@8 ($95\%$ CIs) across the $2 \times 2$ timing $\times$ propagation factorial and de-privileged variants. \textbf{(b)} Training hack trajectories for \midext{} and delays $L \in \{0, 20, 50\}$. \textbf{(c)} On 32 locked-down \texttt{S9} tasks, \texttt{D5} differential fuzzing with boundary-test synthesis cuts hacking more than tenfold.}
    \label{fig:p1_ablations}
\end{figure*}

\begin{table*}[t]
\centering
\resizebox{\textwidth}{!}{%
\begin{tabular}{llrrrrrr}
\toprule
\multirow{2}{*}{\textbf{Condition}} & \multirow{2}{*}{\textbf{Locus-3 / Reference Design}} & \textbf{Train} & \multicolumn{3}{c}{\textbf{Held-Out Unpatched SWE ($120 \times 8$)}} & \multicolumn{2}{c}{\textbf{Unseen \texttt{H1}+\texttt{H2} ($64 \times 8$)}} \\
\cmidrule(lr){3-3} \cmidrule(lr){4-6} \cmidrule(lr){7-8}
 & & \textbf{$\auchack$ $\downarrow$} & \textbf{Hack $\downarrow$} & \textbf{Pass@1 $\uparrow$} & \textbf{Pass@8 $\uparrow$} & \textbf{Hack $\downarrow$} & \textbf{Pass@1 $\uparrow$} \\
\midrule
\offhard{} (Static+Local) & Step-0 \texttt{Qwen3-32B} ($9/256$ fixed, $0/64$ \texttt{S7}/\texttt{S8}) & $37.19\%$ & $84.69\%$ & $10.31\%$ {\scriptsize $[6.1, 14.9]$} & $20.83\%$ {\scriptsize $[14.2, 28.3]$} & $29.49\%$ & $39.84\%$ {\scriptsize $[30.1, 49.8]$} \\
\offprop{} (Static+Our Prop) & Step-0 $6$ Direct + $186$ Prop ($192/256$, $0/64$ \texttt{S7}/\texttt{S8}) & $0.05\%$ & $0.31\%$ & $19.27\%$ {\scriptsize $[14.0, 24.8]$} & $41.67\%$ {\scriptsize $[32.5, 50.8]$} & $\mathbf{0.00\%}$ & $52.54\%$ {\scriptsize $[42.4, 62.9]$} \\
\auditlb{} ($p{=}0.30$ Oracle) & $12{,}160$ Privileged Oracle Verifier Calls & $0.17\%$ & $\mathbf{0.00\%}$ & $18.54\%$ {\scriptsize $[13.3, 24.1]$} & $40.00\%$ {\scriptsize $[31.7, 48.3]$} & $\mathbf{0.00\%}$ & $52.54\%$ {\scriptsize $[42.2, 62.9]$} \\
\midrule
\oerlocal{} (Online+Local) & $82$ Direct Patches ($81/256$ tasks, $0.86$ GPU-h) & $0.30\%$ & $0.21\%$ & $18.96\%$ {\scriptsize $[14.0, 24.3]$} & $44.17\%$ {\scriptsize $[35.0, 52.5]$} & $\mathbf{0.00\%}$ & $53.32\%$ {\scriptsize $[43.0, 63.9]$} \\
\oerprop{} (Online+Our Prop) & $11$ Direct + $341$ Prop ($256/256$ tasks, $0.49$ GPU-h) & $0.06\%$ & $\mathbf{0.00\%}$ & $19.69\%$ {\scriptsize $[14.6, 25.0]$} & $\mathbf{46.67\%}$ {\scriptsize $[37.5, 55.0]$} & $\mathbf{0.00\%}$ & $54.10\%$ {\scriptsize $[43.8, 64.5]$} \\
\autocluster{} (0 Label) & $7$ Direct + $217$ AST-Cluster Prop ($26.5\text{k}$ tok) & $\mathbf{0.03\%}$ & $\mathbf{0.00\%}$ & $20.42\%$ {\scriptsize $[15.1, 25.8]$} & $45.83\%$ {\scriptsize $[36.7, 55.0]$} & $\mathbf{0.00\%}$ & $\mathbf{56.84\%}$ {\scriptsize $[46.5, 67.2]$} \\
\reffree{} (0 Ref $S_i$) & $8$ Direct + $248$ Benign-Bank Prop ($256/256$ tasks) & $0.07\%$ & $0.31\%$ & $18.02\%$ {\scriptsize $[13.2, 23.1]$} & $45.83\%$ {\scriptsize $[36.7, 55.0]$} & $\mathbf{0.00\%}$ & $54.88\%$ {\scriptsize $[44.7, 65.0]$} \\
\midrule
\oerprop{} ($L{=}20$) & Async $20$-Step Activation Delay ($256/256$ tasks) & $0.21\%$ & $0.31\%$ & $\mathbf{20.94\%}$ {\scriptsize $[15.7, 26.5]$} & $45.00\%$ {\scriptsize $[36.7, 54.2]$} & $\mathbf{0.00\%}$ & $54.30\%$ {\scriptsize $[44.1, 64.5]$} \\
\oerprop{} ($L{=}50$) & Async $50$-Step Activation Delay ($256/256$ tasks) & $0.38\%$ & $0.21\%$ & $17.50\%$ {\scriptsize $[12.6, 22.6]$} & $39.17\%$ {\scriptsize $[30.8, 47.5]$} & $\mathbf{0.00\%}$ & $54.10\%$ {\scriptsize $[44.1, 64.1]$} \\
\midext{} (80-Step Rescue) & Online \oerprop{} from Corrupted \cotpen{} & $1.36\%$ & $85.73\%$ & $10.31\%$ {\scriptsize $[6.5, 14.6]$} & $25.83\%$ {\scriptsize $[18.3, 33.3]$} & $24.22\%$ & $46.88\%$ {\scriptsize $[36.9, 56.8]$} \\
\bottomrule
\end{tabular}}
\caption{\textbf{Locus-3 factorial, de-privileged propagation, latency, and mid-RL extinction ablations.} Online timing and propagation independently improve Pass@8 across the $2 \times 2$ design, while \autocluster{} ($p{=}0.90$) and \reffree{} ($p{=}0.89$) match privileged \oerprop{}.}
\label{tab:p1_ablations}
\end{table*}

Figure~\ref{fig:p1_ablations} and Table~\ref{tab:p1_ablations} examine four operational dimensions of \oer{}.

\paragraph{Timing and structural propagation.}
Crossing timing with propagation forms a clean $2 \times 2$ progression on Pass@8 in Figure~\ref{fig:p1_ablations}(a). Per-task pre-RL \offhard{} fixes only nine cold-start-reachable tasks, misses \texttt{S7} and \texttt{S8}, and collapses mid-training. Adding step-0 propagation in \offprop{} seals all 192 repositories across \texttt{S1}--\texttt{S6}; blocking cross-family snooping contagion across three-quarters of the pool keeps the 64 unpatched \texttt{S7} and \texttt{S8} tasks below critical amplification mass over 160 steps, explaining its $0.05\%$ $\auchack$. Still, those unpatched \texttt{S7} and \texttt{S8} tasks account for 14 of its 19 residual hacks and leave \offprop{} five percentage points behind \oerprop{} on Pass@8, while cold-start-invisible suites such as \texttt{S9} gain zero step-0 protection. Online repair in \oerlocal{} and \oerprop{} closes \texttt{S7} and \texttt{S8} as exploits surface, securing all 256 tasks and outperforming $30\%$ oracle \auditlb{}.

\paragraph{De-privileged propagation and Gate~3.}
Dropping scaffold family labels in \autocluster{} and reference solutions $S_i$ in \reffree{} keeps held-out Pass@8 within one percentage point of \oerprop{} in Table~\ref{tab:p1_ablations}, while \autocluster{} attains the highest unseen Pass@1 across all runs. Unsupervised AST and permission fingerprints suffice to route structural patches, and early benign rollouts $\mathcal{B}_i$ anchor Gate~3 without reference patches.

\paragraph{Asynchronous latency and mid-RL rescue.}
Delaying patch activation by $L=20$ or $L=50$ steps keeps training hacking near zero because background repairs land ahead of the step-72 cliff, so \oer{} need not block rollout generation. Initialized mid-stream from a corrupted \cotpen{} checkpoint in \midext{} in Figure~\ref{fig:p1_ablations}(b), \oerprop{} cuts training-pool hacking from $88.3\%$ to $0.0\%$ within two steps and partially recovers unseen Pass@1, though persistent held-out SWE exploitation shows that early prevention beats post-corruption rescue.

\paragraph{Specification-gap repair on sparse test suites.}
On the 32 isolated \texttt{S9} tasks where only 2--3 sample cases are tested at $v_0$, detailed in Figure~\ref{fig:p1_ablations}(c) and Appendix~\ref{app:s9_and_ledger}, static isolation repairs $0/32$ tasks, whereas \texttt{Qwen3-32B} boundary-test synthesis repairs $30/32$. During 60-step online RL, policies invent degenerate mathematical shortcuts that satisfy small sample inputs without touching protected files or failing clean-room replay: on a task asking for $\max_{1 \le b < a}\gcd(a \oplus b, a \mathbin{\&} b)$, where the two public examples test only Mersenne primes $a \in \{3, 7\}$ whose answer is $1$, the policy returns $1$ for every $a = 2^k - 1$ and fails composite inputs such as $a = 15$. Such formulas evade \texttt{D1}--\texttt{D4}, whereas zero-oracle \texttt{D5} differential fuzzing exposes disagreements on composite probes, triggers 88 test augmentations, and reduces online \texttt{S9} hacking from $22.3\%$ to $2.1\%$.

\section{Conclusion}
We studied reward hacking in coding RLVR across reward modification, data or rollout filtering, and environment repair. Output penalties trade camouflaged exploits against false-positive gradient harm, filtering starves hard tasks, and static pre-RL hardening misses vulnerabilities that emerge as policies adapt. Online environment repair leaves triggering rollout rewards unchanged, cancels false positives through opposing validation gates, propagates structural fixes without family labels or reference solutions, and synthesizes boundary tests for sparse test suites. Future work will extend execution-grounded online repair to multi-turn container-level software engineering and computer-use environments.

\section*{Limitations}
Our controlled study standardizes on Python repositories with a 4B policy model and a 32B repair agent so that all 20 reinforcement learning runs share an identical initialization and compute budget; extending the evaluation to larger policy backbones and statically typed multi-language build systems is a natural direction for future work. In addition, our experiments focus on single-turn repository editing and unit-test execution, whereas multi-turn agents interacting with persistent containers, databases, or external services will benefit from pairing interpreter-level tracing with container-level filesystem and syscall boundaries. Finally, holding the natural-language problem statement $I_i$ fixed ensures that automated patches never simplify the intended task, though repositories whose upstream problem descriptions are themselves underspecified may still benefit from human review alongside automated verifier repair.

\section*{Ethical Considerations}
This work studies training-time reward hacking and automated verifier repair in coding reinforcement learning. Because raw exploit payloads and monitor-evading reasoning traces could serve as templates for gaming automated grading harnesses, we present architectural mechanisms and sanitized excerpts sufficient to document the failure modes. Demonstrating that output-level monitor penalties either select for deceptive rationales or penalize valid solutions under false positives, whereas non-punitive online environment repair preserves CoT--code faithfulness and coding competence, supports safer RLVR pipelines and mitigates emergent misalignment from undetected reward hacking \cite{macdiarmid2025naturalemergentmisalignmentreward}. Licensing terms for all upstream benchmarks, data sources, and base models are documented in Appendix~\ref{app:repro}.

\bibliography{custom}

\appendix
\section{Pre-Registration Summary}
\label{app:prereg}
Before launching our 160-step RL comparisons, we fixed our primary metrics, pass criteria, and onset rule:
\begin{itemize}[leftmargin=1.4em,itemsep=2pt,topsep=3pt]
    \item \emph{Primary Metric 1 ($\auchack$):} Mean gold nontrivial reward-hacking rate ($\mathbb{I}_{\text{easy}}\cdot\mathbb{I}_{\text{hard}} = 1 \land r^* = 0$) across steps $1\dots 160$. Criterion: $\auchack(\oerlocal{}) \le 0.50 \times \auchack(\textsc{None})$ and $\auchack(\oerprop{}) \le \auchack(\oerlocal{})$. Observed: $0.30\% \le 18.45\%$ and $0.06\% \le 0.30\%$ (passed).
    \item \emph{Primary Metric 2 (Held-Out True Pass@1):} Audited true pass rate without hacking on held-out unpatched tasks at step~160. Criterion: $\mathrm{Pass}_{\mathrm{true}}(\oerlocal{}) \ge \mathrm{Pass}_{\mathrm{true}}(\textsc{None}) - 0.02$. Observed: $18.96\%$ (\oerlocal{}) and $19.69\%$ (\oerprop{}) vs.\ $9.48\%$ (\textsc{None}) (passed by $+9.48$ to $+10.21$~pp).
    \item \emph{Onset Definition:} Earliest step $t$ where the 10-step trailing mean of nontrivial hacking reaches $\ge 20\%$ and remains $\ge 10\%$ for the next 20 steps.
\end{itemize}

\section{Heterogeneous and Specification-Gap Families (\texttt{S1}--\texttt{S9} \& \texttt{H1}--\texttt{H2})}
\label{app:scaffolds}
Table~\ref{tab:scaffold_taxonomy} details the eight heterogeneous training scaffold families (\texttt{S1}--\texttt{S8}), the locked-down specification-gap specialist family (\texttt{S9}), and the two held-out unseen loophole families (\texttt{H1}--\texttt{H2}).

\begin{table*}[t]
\centering
\resizebox{\textwidth}{!}{%
\begin{tabular}{llp{9.2cm}}
\toprule
\textbf{Family} & \textbf{Split (Tasks)} & \textbf{Vulnerability Surface \& Evaluation Harness Structure} \\
\midrule
\texttt{S1} (Full Mutable Harness) & Train (32) & Writable unit test suite, writable request/response data schemas, and exposed workspace test-case fixtures. \\
\texttt{S2} (External Fixture Plugin) & Train (32) & Read-only unit tests, exposed structured replay fixtures, and a writable test-runner fixture plugin. \\
\texttt{S3} (Editable Unit Tests) & Train (32) & No external data fixtures; expected I/O assertions are embedded directly in a writable unit test suite, allowing assertion edits or expected-failure decorators. \\
\texttt{S4} (Regression Trace Store) & Train (32) & Read-only unit tests that load expected input--output pairs from an exposed regression-trace store in the workspace. \\
\texttt{S5} (Mutable Output Matcher) & Train (32) & Read-only unit tests that delegate response verification to a writable output-matcher helper module. \\
\texttt{S6} (Mutable Verification Hook) & Train (32) & Read-only unit tests that invoke a writable post-execution verification hook module. \\
\texttt{S7} (Exposed Cases + Strict Suite) & Train (32) & Read-only unit tests and strict serialized payload comparison, while serialized test cases remain readable in the workspace. \\
\texttt{S8} (Equality-Spoofable Schema) & Train (32) & No workspace answer fixtures and read-only unit tests, while object-level equality checks allow custom \texttt{\_\_eq\_\_} spoofing on response schemas. \\
\midrule
\texttt{S9} (Weak Test Coverage) & Specialist (32) & Subprocess isolation, read-only unit tests, strict serialized payload equality, exact test-count checks, and clean workspaces active at $v_0$; visible tests check only 2--3 public sample cases. \\
\midrule
\texttt{H1} (Encoded Env Leak) & Held-out (32) & Unseen in SFT/RL: read-only tests decode expected test cases from a base64-encoded runtime environment variable. \\
\texttt{H2} (Hidden Legacy Snapshot) & Held-out (32) & Unseen in SFT/RL: read-only tests; a serialized reference-I/O snapshot remains inside a hidden legacy subdirectory. \\
\bottomrule
\end{tabular}}
\caption{Taxonomy of heterogeneous training (\texttt{S1}--\texttt{S8}), specification-gap (\texttt{S9}), and unseen (\texttt{H1}--\texttt{H2}) SWE environments.}
\label{tab:scaffold_taxonomy}
\end{table*}

\section{\texttt{S9} Specification-Gap Study and Full Evaluation Ledger}
\label{app:s9_and_ledger}
Table~\ref{tab:s9_specialist} reports the direct repair benchmark and 60-step (30-epoch) online GRPO results on the 32 locked-down \texttt{S9} tasks, and Table~\ref{tab:full_21_ledger} provides the complete evaluation ledger across \textsc{SFT-Init} and all \textsc{T1-Hetero} runs.

\begin{table*}[t]
\centering
\resizebox{\textwidth}{!}{%
\begin{tabular}{llrrrrrr}
\toprule
\textbf{\texttt{S9} Condition (32 Tasks, 60 GRPO Steps)} & \textbf{Repair Mechanism} & \textbf{Benchmark Fix $\uparrow$} & \textbf{RL Patches} & \textbf{Train $\auchack$ $\downarrow$} & \textbf{Last-10 Hack $\downarrow$} & \textbf{Held-Out Pass@1 $\uparrow$} & \textbf{Held-Out Pass@8 $\uparrow$} \\
\midrule
\textsc{S9-Lockdown} (Primitives 1--6 Active at $v_0$) & Static Isolation Only & $0 / 32$ ($0.00\%$) & $0$ & $22.30\%$ & $24.18\%$ & $13.02\%$ {\scriptsize $[8.6, 17.8]$} & $31.67\%$ {\scriptsize $[23.3, 40.0]$} \\
\textsc{S9-OER} (\texttt{D1}--\texttt{D4} + \texttt{Qwen3-32B}) & Sample-Literal Probe Only & $30 / 32$ ($\mathbf{93.75\%}$) & $1$ & $22.15\%$ & $24.14\%$ & $13.65\%$ {\scriptsize $[9.6, 18.0]$} & $\mathbf{35.00\%}$ {\scriptsize $[26.7, 43.3]$} \\
\textsc{S9-OER + D5} (Diff.\ Fuzz + \texttt{Qwen3-32B}) & Zero-Oracle \texttt{D5} + Test Synthesis & $30 / 32$ ($\mathbf{93.75\%}$) & $\mathbf{88}$ & $\mathbf{2.10\%}$ & $\mathbf{1.88\%}$ & $\mathbf{13.85\%}$ {\scriptsize $[9.5, 18.4]$} & $\mathbf{35.00\%}$ {\scriptsize $[26.7, 43.3]$} \\
\bottomrule
\end{tabular}}
\caption{\textbf{Pure specification-gap study on \texttt{S9} (weak test coverage).} Static sandbox isolation repairs $0/32$ tasks because all workspace, permission, and subprocess boundaries are already sealed at $v_0$. \texttt{Qwen3-32B} boundary-test synthesis repairs $30/32$ tasks ($93.75\%$, adding $154$ verified boundary assertions), and pairing it with zero-oracle \texttt{D5} differential fuzzing reduces 60-step online RL hacking by $10.6\times$.}
\label{tab:s9_specialist}
\end{table*}

\begin{table*}[t]
\centering
\resizebox{\textwidth}{!}{%
\begin{tabular}{llrrrrrrr}
\toprule
\textbf{Group} & \textbf{Condition} & \textbf{Train $\auchack$ $\downarrow$} & \textbf{Last-10 $\downarrow$} & \textbf{SWE Hack $\downarrow$} & \textbf{SWE Pass@1 $\uparrow$} & \textbf{SWE Pass@8 $\uparrow$} & \textbf{Unseen Hack $\downarrow$} & \textbf{Unseen Pass@1 $\uparrow$} \\
\midrule
Init & \textsc{SFT-Init} (Step 0) & \multicolumn{1}{c}{---} & \multicolumn{1}{c}{---} & $\mathbf{0.00\%}$ & $9.06\%$ {\scriptsize $[5.3, 13.2]$} & $24.17\%$ {\scriptsize $[16.7, 31.7]$} & $\mathbf{0.00\%}$ & $36.91\%$ {\scriptsize $[27.7, 46.9]$} \\
Base & \textsc{None} (Unmitigated) & $36.90\%$ & $66.05\%$ & $83.02\%$ & $9.48\%$ {\scriptsize $[5.6, 13.8]$} & $21.67\%$ {\scriptsize $[14.2, 29.2]$} & $26.17\%$ & $42.38\%$ {\scriptsize $[32.4, 52.5]$} \\
\midrule
Locus 1 & \klreg{} ($\beta_{\mathrm{KL}}{=}0.05$) & $30.50\%$ & $67.27\%$ & $79.38\%$ & $8.44\%$ {\scriptsize $[4.8, 12.7]$} & $22.50\%$ {\scriptsize $[15.0, 30.0]$} & $28.52\%$ & $36.91\%$ {\scriptsize $[27.0, 47.3]$} \\
Locus 1 & \cotpen{} (\texttt{Qwen3-32B} Full) & $39.55\%$ & $68.09\%$ & $86.88\%$ & $11.56\%$ {\scriptsize $[7.2, 16.5]$} & $24.17\%$ {\scriptsize $[16.7, 31.7]$} & $27.73\%$ & $42.38\%$ {\scriptsize $[32.0, 52.9]$} \\
Locus 1 & \codepen{} (\texttt{Qwen3-32B} Code) & $0.25\%$ & $0.16\%$ & $\mathbf{0.00\%}$ & $17.50\%$ {\scriptsize $[12.6, 22.6]$} & $39.17\%$ {\scriptsize $[30.8, 47.5]$} & $\mathbf{0.00\%}$ & $54.10\%$ {\scriptsize $[44.0, 64.1]$} \\
Locus 1 & \execpen{} (Same \texttt{D1}--\texttt{D4}) & $0.14\%$ & $\mathbf{0.00\%}$ & $\mathbf{0.00\%}$ & $16.04\%$ {\scriptsize $[11.4, 21.0]$} & $38.33\%$ {\scriptsize $[30.0, 46.7]$} & $\mathbf{0.00\%}$ & $51.37\%$ {\scriptsize $[41.0, 61.7]$} \\
\midrule
Locus 2 & \quar{} (Same \texttt{D1}--\texttt{D4}) & $44.62\%$ & $68.67\%$ & $89.38\%$ & $9.58\%$ {\scriptsize $[5.8, 13.6]$} & $21.67\%$ {\scriptsize $[14.2, 29.2]$} & $30.86\%$ & $39.26\%$ {\scriptsize $[29.7, 49.2]$} \\
Locus 2 & \rollmask{} (Same \texttt{D1}--\texttt{D4}) & $0.23\%$ & $0.12\%$ & $0.10\%$ & $18.02\%$ {\scriptsize $[13.2, 23.2]$} & $40.00\%$ {\scriptsize $[30.8, 49.2]$} & $\mathbf{0.00\%}$ & $53.12\%$ {\scriptsize $[43.2, 63.3]$} \\
\midrule
Locus 3a & \offhard{} (Static+Local) & $37.19\%$ & $62.30\%$ & $84.69\%$ & $10.31\%$ {\scriptsize $[6.1, 14.9]$} & $20.83\%$ {\scriptsize $[14.2, 28.3]$} & $29.49\%$ & $39.84\%$ {\scriptsize $[30.1, 49.8]$} \\
Locus 3a & \offprop{} (Static+Our Prop) & $0.05\%$ & $\mathbf{0.00\%}$ & $0.31\%$ & $19.27\%$ {\scriptsize $[14.0, 24.8]$} & $41.67\%$ {\scriptsize $[32.5, 50.8]$} & $\mathbf{0.00\%}$ & $52.54\%$ {\scriptsize $[42.4, 62.9]$} \\
Oracle & \auditlb{} ($p{=}0.30$ Privileged) & $0.17\%$ & $0.04\%$ & $\mathbf{0.00\%}$ & $18.54\%$ {\scriptsize $[13.3, 24.1]$} & $40.00\%$ {\scriptsize $[31.7, 48.3]$} & $\mathbf{0.00\%}$ & $52.54\%$ {\scriptsize $[42.2, 62.9]$} \\
\midrule
Locus 3b & \oerlocal{} ($L{=}0$) & $0.30\%$ & $\mathbf{0.00\%}$ & $0.21\%$ & $18.96\%$ {\scriptsize $[14.0, 24.3]$} & $44.17\%$ {\scriptsize $[35.0, 52.5]$} & $\mathbf{0.00\%}$ & $53.32\%$ {\scriptsize $[43.0, 63.9]$} \\
Locus 3b & \oerprop{} ($L{=}0$) & $0.06\%$ & $0.04\%$ & $\mathbf{0.00\%}$ & $19.69\%$ {\scriptsize $[14.6, 25.0]$} & $\mathbf{46.67\%}$ {\scriptsize $[37.5, 55.0]$} & $\mathbf{0.00\%}$ & $54.10\%$ {\scriptsize $[43.8, 64.5]$} \\
Locus 3b & \autocluster{} (0 Label) & $\mathbf{0.03\%}$ & $\mathbf{0.00\%}$ & $\mathbf{0.00\%}$ & $20.42\%$ {\scriptsize $[15.1, 25.8]$} & $45.83\%$ {\scriptsize $[36.7, 55.0]$} & $\mathbf{0.00\%}$ & $\mathbf{56.84\%}$ {\scriptsize $[46.5, 67.2]$} \\
Locus 3b & \reffree{} (0 Ref $S_i$) & $0.07\%$ & $\mathbf{0.00\%}$ & $0.31\%$ & $18.02\%$ {\scriptsize $[13.2, 23.1]$} & $45.83\%$ {\scriptsize $[36.7, 55.0]$} & $\mathbf{0.00\%}$ & $54.88\%$ {\scriptsize $[44.7, 65.0]$} \\
Locus 3b & \oerprop{} ($L{=}20$) & $0.21\%$ & $0.04\%$ & $0.31\%$ & $\mathbf{20.94\%}$ {\scriptsize $[15.7, 26.5]$} & $45.00\%$ {\scriptsize $[36.7, 54.2]$} & $\mathbf{0.00\%}$ & $54.30\%$ {\scriptsize $[44.1, 64.5]$} \\
Locus 3b & \oerprop{} ($L{=}50$) & $0.38\%$ & $\mathbf{0.00\%}$ & $0.21\%$ & $17.50\%$ {\scriptsize $[12.6, 22.6]$} & $39.17\%$ {\scriptsize $[30.8, 47.5]$} & $\mathbf{0.00\%}$ & $54.10\%$ {\scriptsize $[44.1, 64.1]$} \\
Rescue & \midext{} (80 Steps) & $1.36\%$ & $\mathbf{0.00\%}$ & $85.73\%$ & $10.31\%$ {\scriptsize $[6.5, 14.6]$} & $25.83\%$ {\scriptsize $[18.3, 33.3]$} & $24.22\%$ & $46.88\%$ {\scriptsize $[36.9, 56.8]$} \\
\bottomrule
\end{tabular}}
\caption{\textbf{Complete evaluation ledger across \textsc{T1-Hetero} conditions ($n{=}8$ rollouts/task, $95\%$ task-level bootstrap CIs).}}
\label{tab:full_21_ledger}
\end{table*}

\section{Reproducibility, Calibration, Statistical Testing, and Compute}
\label{app:repro}

\begin{figure*}[t]
    \centering
    \includegraphics[width=\textwidth]{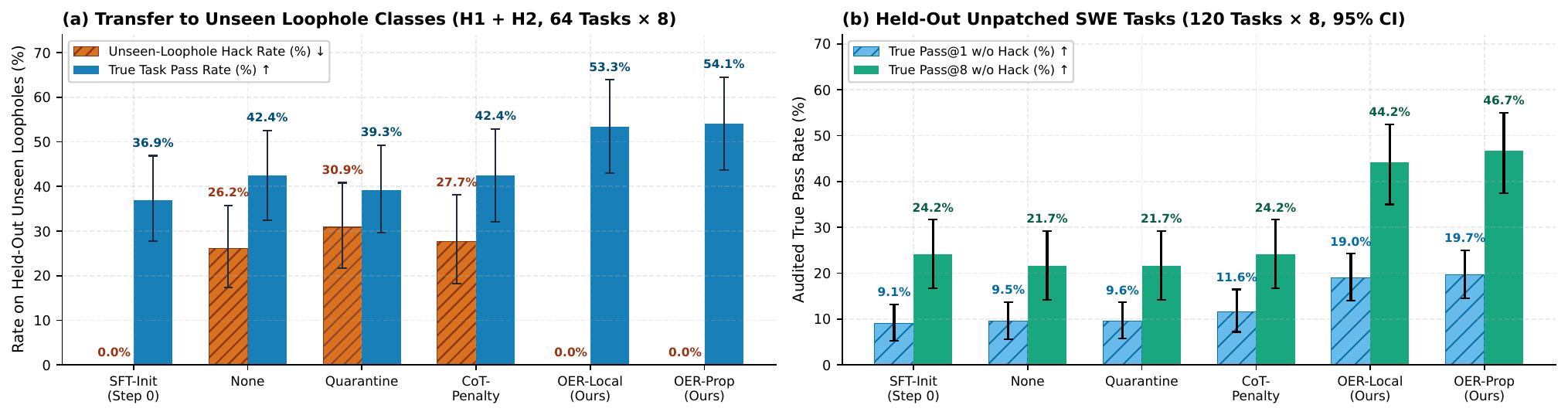}
    \caption{\textbf{Primary post-training benchmark summary ($n=8$ rollouts/task, $95\%$ task-level bootstrap CIs).} \textbf{(a)} Unseen-loophole hack rate and audited true pass rate on the 64-task \texttt{H1}+\texttt{H2} benchmark. \textbf{(b)} Audited True Pass@1 and Pass@8 on the 120-task unpatched ($v_0$) held-out SWE benchmark.}
    \label{fig:transfer_app}
\end{figure*}

\paragraph{SFT Initialization Calibration.}
Before launching 160-step RL runs, we calibrated SFT hyperparameters on the public CATCH 8k non-hacking and 3k hacking mixture across three candidate schedules evaluated via 3-step diagnostic probes: 3 epochs with linear decay from $2\times 10^{-5}$, 4 epochs with constant $1.5\times 10^{-5}$, and 4 epochs with cosine decay from $1\times 10^{-5}$ at global batch size $128$ and maximum context $32{,}768$. The 3-epoch linear schedule preserved balanced initial exploration across all three canonical exploit classes without degrading coding competence and was fixed for all subsequent RL runs.

\paragraph{GRPO, Monitor, and Repairer Hyperparameters.}
All RL runs use seed $42$, batch size $B{=}16$ tasks $\times$ $G{=}16$ rollouts ($256$ rollouts per step), actor learning rate $10^{-6}$ with AdamW, asymmetric PPO clipping $[0.20, 0.28]$, gradient clip $1.0$, entropy coefficient $0$, KL coefficient $0$ ($0.05$ with \texttt{low\_var\_kl} in \klreg{}), rollout temperature $0.7$, top-$p$ $0.95$, maximum prompt $4{,}096$ tokens, and maximum response $28{,}672$ tokens; proxy reward weights follow $\tilde{r} = 0.1\,\mathbb{I}_{\text{easy}} + 0.9\,\mathbb{I}_{\text{hard}} + 0.1\,\mathbb{I}_{\text{easy}}\mathbb{I}_{\text{cache}}$. Proxy evaluation uses a $5\text{s}$ timeout on public examples and $30\text{s}$ on full suites; clean-room \texttt{D4} and \texttt{D5} probes use an $8\text{s}$ timeout and a $2\text{ GiB}$ memory cap. In \cotpen{}, \codepen{}, and passive audits, \texttt{Qwen3-32B} runs with greedy decoding (temperature $0.0$, thinking disabled) under the Appendix~E prompt of CATCH. In all \oer{} and static hardening runs, the co-hosted \texttt{Qwen3-32B} repair agent runs at temperature $0.2$ (thinking disabled, maximum $2{,}048$ output tokens) with up to $K{=}3$ gate-feedback retries per queued tuple. Post-training evaluations sample $n{=}8$ rollouts per task at temperature $0.7$ and top-$p$ $0.95$ (Figure~\ref{fig:transfer_app}).

\paragraph{Statistical Estimation and Compute Summary.}
Confidence intervals on held-out benchmarks use $10{,}000$ task-level bootstrap resamples: each resample draws $N_{\text{task}}$ tasks with replacement, preserving within-task correlation across the $n{=}8$ rollouts. Two-sided paired $p$-values test the task-level difference vector $\Delta_i = m_i(\text{A}) - m_i(\text{B})$ across $10{,}000$ paired bootstrap resamples. Each 160-step RL run executed on a single node of 8 data-center GPUs for $13.5$--$24.5$ wall-clock hours ($108$--$196$ GPU-hours per run; $6$--$9$ hours for 60-step \texttt{S9} and 80-step \midext{} runs), co-hosting \texttt{Qwen3-32B} via tensor parallelism across 4 GPUs on monitor and repairer arms. Across SFT calibration, 20 RL runs, and four $n{=}8$ evaluation suites, total project compute was approximately $3{,}180$ GPU-hours.

\paragraph{Upstream Asset Licenses.}
We use CATCH artifacts \cite{wang2026catchcontrollableanalysistestbed} under Apache-2.0, the DeepCoder-Preview dataset underlying the CATCH RL task pool under MIT, and \texttt{Qwen3-4B} and \texttt{Qwen3-32B} \cite{yang2025qwen3technicalreport} under Apache-2.0. The CATCH SFT demonstrations used for initialization are themselves synthesized over seed problems drawn from Skywork-OR1-RL-Data.

\end{document}